\documentclass[conference,a4paper]{IEEEtran}
\IEEEoverridecommandlockouts
\IEEEpubid{\makebox[\columnwidth]
{\hfill }
\hspace{\columnsep}\makebox[\columnwidth]{ }}
\usepackage{cite}
\usepackage{amssymb,amsfonts}
\usepackage{algorithmic}
\usepackage{graphicx}
\usepackage{epsfig}
\usepackage{mathptmx}
\usepackage{amssymb}
\usepackage{textcomp}
\usepackage{siunitx}
\usepackage{xcolor}
\usepackage[cmex10]{amsmath}
\usepackage{booktabs}
\usepackage{multirow}
\usepackage{caption}
\usepackage{subcaption}
\usepackage{float}
\def\BibTeX{{\rm B\kern-.05em{\sc i\kern-.025em b}\kern-.08em
    T\kern-.1667em\lower.7ex\hbox{E}\kern-.125emX}}
    
    \makeatletter
 \let\old@ps@headings\ps@headings
 \let\old@ps@IEEEtitlepagestyle\ps@IEEEtitlepagestyle
 \def\confheader#1{%
 \def\ps@headings{%
 \old@ps@headings%
 \def\@oddhead{\strut\hfill#1\hfill\strut}%
 \def\@evenhead{\strut\hfill#1\hfill\strut}%
 }%
 \def\ps@IEEEtitlepagestyle{%
 \old@ps@IEEEtitlepagestyle%
 \def\@oddhead{\strut\hfill#1\hfill\strut}%
 \def\@evenhead{\strut\hfill#1\hfill\strut}%
 }%
 \ps@headings%
 }
\makeatother

\confheader{}

\graphicspath{{./figures/}} 
\begin{document}

\title{Adaptive Acoustic Flow-Based Navigation\\ with 3D Sonar Sensor Fusion\\
}

\author{\IEEEauthorblockN{Jansen Wouter Laurijssen Dennis and Steckel Jan}
\IEEEauthorblockA{\textit{FTI-CoSys Lab, University of Antwerp} \\
Antwerp, Belgium \\
\textit{Flanders Make Strategic Research Centre}\\
Lommel, Belgium \\
jan.steckel@uantwerpen.be}
}

\maketitle

\begin{abstract}
Navigating spatially varied and dynamic environments is one of the key tasks for autonomous agents. In this paper we present a novel method of navigating a mobile platform with one or multiple 3D-sonar sensors. Moving a mobile platform and subsequently any 3D-sonar sensor on it, will create signature variations over time of the echoed reflections in the sensor readings. An approach is presented to create a predictive model of these signature variations for any motion type. Furthermore, the model is adaptive and works for any position and orientation of one or multiple sonar sensors on a mobile platform. We propose to use this adaptive model and fuse all sensory readings to create a layered control system allowing a mobile platform to perform a set of primitive motions such as collision avoidance, obstacle avoidance, wall following and corridor following behaviours to navigate an environment with dynamically moving objects within it. This paper describes the underlying theoretical base of the entire navigation model and validates it in a simulated environment with results that shows the system is stable and delivers expected behaviour for several tested spatial configurations of one or multiple sonar sensors that can complete an autonomous navigation task.  
\end{abstract}


\section{Introduction}
In recent years, autonomous navigation of robotic mobile platforms has seen enormous improvements in both the sensing and control aspects. Various sensor types such as cameras, laser-scanners and radar have been used to provide the system with  information to create a map of the environment, recognise objects, localise the autonomous agent within that map and simultaneously navigate the environment. However, there are still circumstances where these popular sensory inputs fail. Specifically, optical sensors that depend on the projection of structured light (cameras, laser-scanners) operate sub-optimally when their medium is distorted by for example smoke, dust, airborne particles, mist or changes in illumination depending on the time of day or caused by extreme weather conditions.  \\
In-air sonar has become a subject of research and industrial interest to extract information from the environment to support autonomous navigation beyond simple collision avoidance. However, not the typical sonar sensor found on a car used as a parking-sensor, but an advanced sonar sensor with a wide field of view (FOV), ability to cope with simultaneously arriving echoes and capable of extracting full 3D-spatial information from the reflections in the environments. In recent years, such biologically inspired \cite{Griffin1974} sensors have been developed by Jan Steckel et al. \cite{Steckel2020, Kerstens2019} and have been used for applications with semi-autonomous agents to Simultaneously Localise and Map (SLAM) \cite{Steckel2013b, Kerstens2020} the environment and navigate it locally\cite{Steckel2017}. However, a common issue still seen with these advanced sonar sensors, is the limited spatial resolution one sonar sensor can deliver of the full spatial environment around for example a mobile platform. Generally, this issue is caused by leakage of the point spread function of the sonar sensor which causes certain reflections not to be picked up by the sensor. Furthermore, these sensors have a FOV of up to \ang{180} and as such cannot sense the entire area around all sides of a mobile platform. To solve this problem, multiple sonar sensors (Multi-Sonar) can be used simultaneously and allow to have a spatial resolution beyond what the sampling resolution of one sensor allows \cite{Gazit2009}. While processing the data of multiple 3D-sonar sensors is computationally expensive, recent work allows for synchronising and processing several 3D sensors simultaneously on a GPU-powered system in real-time \cite{Jansen2020}. This creates a data-stream coming from multiple sensors of acoustic images holding cues of the environment allowing for several new and improved possible applications in comparison to using only one 3D-sonar sensor. For example when they are mounted on a mobile platform such as an Autonomous Ground Vehicle (AGV).\\
Inspired by examples found in nature such as insects using optical (flow) cues \cite{Franceschini2007, Srinivasan2011} or bats using acoustic (flow) cues\cite{Corcoran2017,Simon2020,Greiter2017,Warnecke2016}, research has shown to be able to use these cues to extract information about both the ego-motion of the agent through an environment as well as the spatial structure of that environment through that motion. In a previous publication by Jan Steckel et al. \cite{Steckel2020} these acoustic flow cues were used to create a control algorithm for a mobile platform that allows obstacle avoidance and corridor following behaviour through hallway environments featuring corners and T-sections. However, it had a major restriction in the configuration of the mobile platform: only one sonar sensor could be used and had to be aligned and placed closely to the centre of the mobile platform. 
\newpage
\noindent This is sub-optimal as most real systems used in industrial applications will not allow for the mounting of a single sensor in the centre of the vehicle and perceive the total area around it. In this paper we use the benefits of operating multiple 3D-sonar sensors in real-time simultaneously to increase the sensor sensing space creating sensor fusion. Furthermore, we have implemented a heavily improved control system in a modular way to allow for any sensor configuration so that they can adhere to the potential placement constraints of the vehicle and to continue to operate with stable and expected behaviours. In this paper we establish the underlying theoretical foundation for this system and perform a numerical evaluation in simulation. 
\section{eRTIS: Embedded Real-Time Imaging Sonar }\label{sec:ertis}
\begin{figure}
\centering
\includegraphics[width=0.5\linewidth]{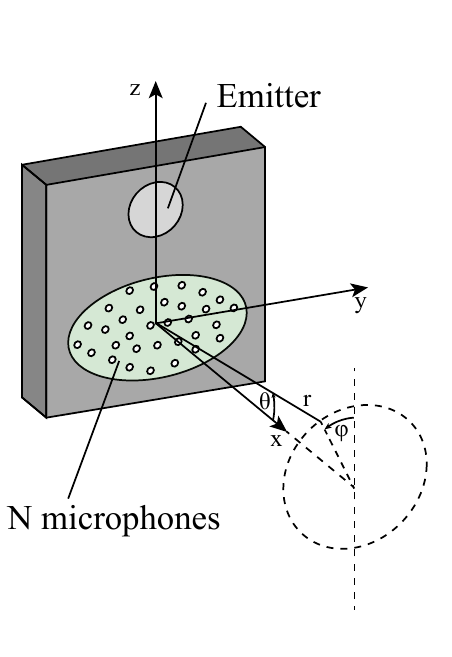}
\caption{Position of reflector ($r,\theta,\varphi$) expressed in the spherical coordinate system associated with the eRTIS sensor.}
\label{fig:ertis_coordinates}
\end{figure}

We will provide a brief overview of the sonar sensor used, the Embedded Real-Time Imaging Sonar (eRTIS) \cite{Steckel2020}. An in-depth article on eRTIS can be found in \cite{Kerstens2019}. eRTIS has a single emitter and uses a known, irregularly positioned array of 32 digital microphones. The emitter will send out a broadband FM-chirp of which its reflections will be captured by the microphones when it is reflected on objects. Using these specific chirps, we can increase the spatial resolution \cite{Steckel2013}. Thereupon, the recorded microphone signals are processed to create the spatial images in a spherical coordinate system with ($r,\theta,\varphi$) as shown in Fig. \ref{fig:ertis_coordinates}. We call these images Energyscapes. eRTIS sensors have a horizontal and vertical FOV of \ang{180}. We recently created a system architecture and processing algorithm where we synchronise multiple eRTIS sensors and process them in real-time \cite{Jansen2020}. This resulted in allowing a mobile platform equipped with an NVIDIA Jetson TX2 to use three eRTIS sensors and process their 2D images in real-time at a measurement frequency of \SI{10}{\hertz}. We target similar systems in this paper with a simulated mobile platform with up to three synchronised eRTIS sensors.
\section{Acoustic Flow Model}\label{sec:acoustic_flow_model}
\begin{figure}
\centering
\includegraphics[width=1\linewidth]{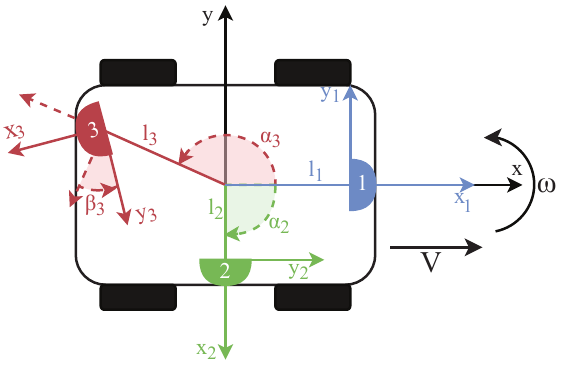}
\caption{Top view of a mobile platform with 3 sonar sensors attached. The sensors are spatially defined by the distance $l$ from the mobile platform's centre, the angle $\alpha$ around on the XY-plane of the mobile platform relative to the x-axis and lastly the angle $\beta$ defining the local rotation of the sensor around its pivot.}
\label{fig:robot_coordinates}
\end{figure}

The sensory readings of the eRTIS sensor directly measures the coordinates ($r,\theta,\varphi$) of all reflecting objects in the environment. The model we propose contains a set of differential equations that describe the perceived change of the origin of these reflections in 3D space with the linear and rotational components of the ego-motion of the mobile platform.\\ Furthermore, we will look into two particular cases where we construct the 2D-velocity field specifically for only linear movement and only rotation movement. The previous acoustic flow model on which this paper is based is described in \cite{Steckel2017} for a single forward facing sonar sensor. In this section we will focus on the changes made to support a Multi-Sonar configuration where a sensor can be placed anywhere on a mobile platform with the only restrictions that they all lie on the same horizontal plane. An example of a possible Multi-Sonar configuration can be seen in Fig. \ref{fig:robot_coordinates}.

\begin{figure*}[!ht]
\centering
  \begin{tabular}[b]{cc}
    \begin{subfigure}[b]{0.30\linewidth}
      \includegraphics[width=\textwidth]{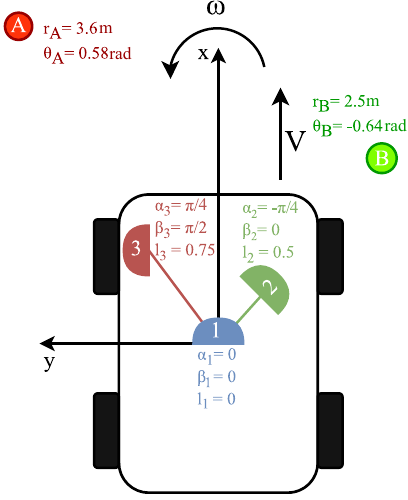}
      \caption{}
      \label{fig:af_flow_lines_robot_config}
    \end{subfigure}
    &
    \begin{tabular}[b]{c}
      \begin{subfigure}[b]{0.65\linewidth}
        \includegraphics[width=\textwidth]{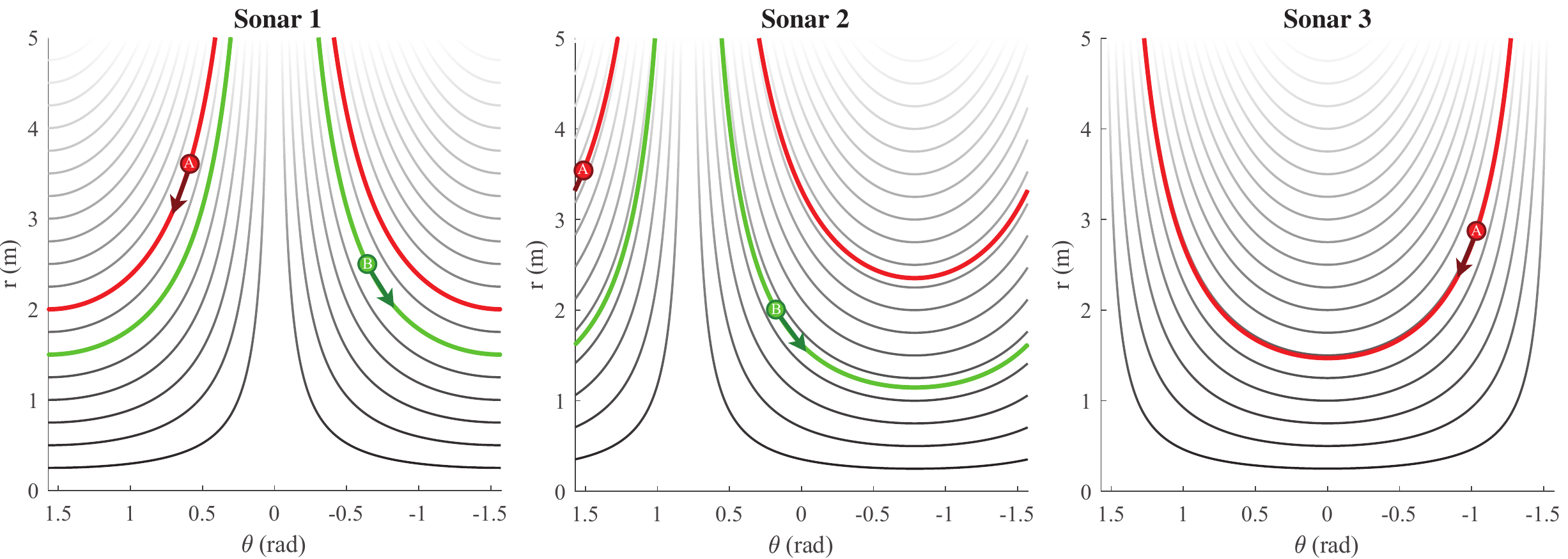}
        \caption{}
        \label{fig:af_flow_lines_linear}
      \end{subfigure}\\
      \begin{subfigure}[b]{0.65\linewidth}
        \includegraphics[width=\textwidth]{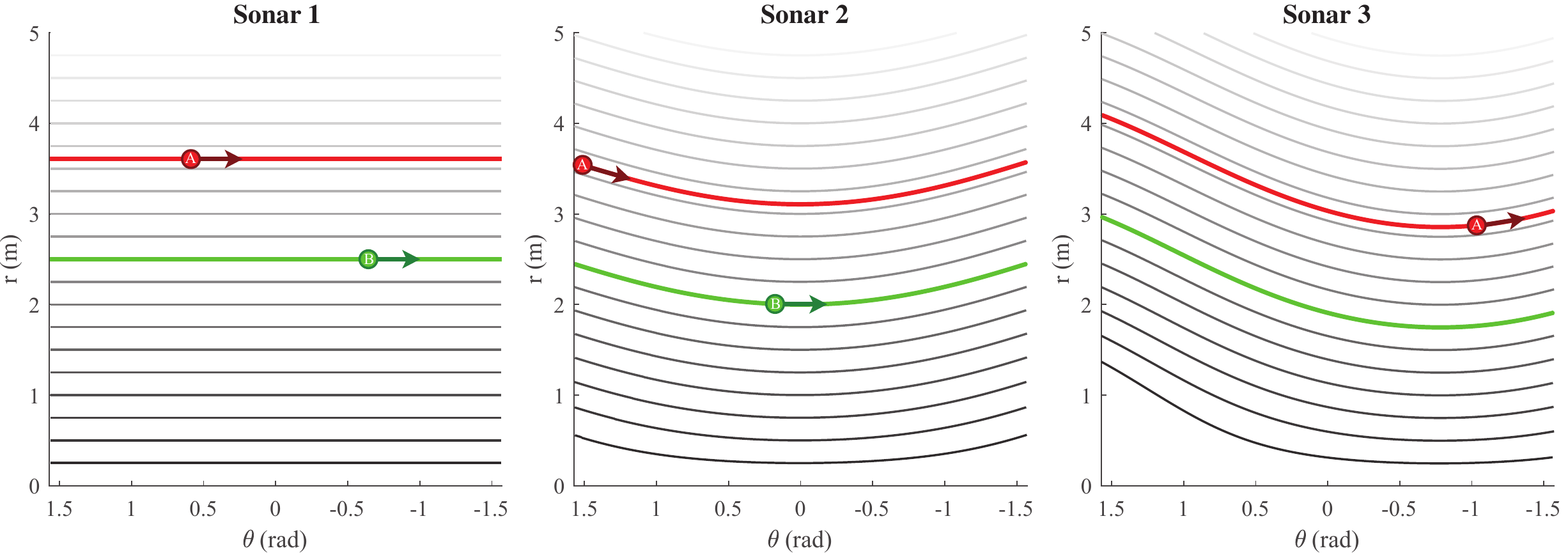}
        \caption{}
        \label{fig:af_flow_lines_rotation}
      \end{subfigure}
    \end{tabular}
  \end{tabular}
  \label{fig:af_flow_lines}
  \caption{(a) An example mobile platform (top view, illustrated scale) with Multi-Sonar defined as 1, 2 and 3. The $\alpha,\beta$(radians) and $l$(meters) values are listed next to them. Two reflectors are shown defined as A and B with their $(r_0,\theta_0$) coordinates listed next to them; (b) The $(r(t),\theta(t))$ trajectories (=flow-lines) corresponding with reflectors A and B for a linear movement are highlighted for each sonar of Fig. \ref{fig:af_flow_lines_robot_config}. We also show the trajectories of possible reflectors with distances falling in range ($r \in [0; r_{max}]$) with $r_{max}$ the maximum range of the sonar sensor. The arrows of the reflectors show the direction of their trajectory if a positive linear movement forward ($V$) would occur; (c) The flow-lines corresponding with the reflectors A and B for a rotation movement are highlighted for each sonar of Fig. \ref{fig:af_flow_lines_robot_config}. The trajectories of possible reflectors with distances falling in range ($r \in [0; r_{max}]$) are shown as well. The arrows of the reflectors show the direction of their trajectory if a positive rotation movement ($\omega$) would occur.}
\end{figure*}
\subsection{2D-velocity field}
First we will derive the general 2D-velocity field equations. We use a spherical coordinate system to define the position of a reflector in sonar sensor reference frame by the vector $(r,\theta,\varphi)$ as shown in Fig. \ref{fig:ertis_coordinates}. 
This coordinate system is related to the standard right-handed Cartesian coordinates by
\begin{equation}
\vec{x}(t) =
\begin{bmatrix}
    x(t)  \\
    y(t)  \\
    z(t)
\end{bmatrix}=
\begin{bmatrix}
    r(t)\cos(\theta(t))  \\
    -r(t)\sin(\theta(t))\sin(\varphi(t))  \\
    r(t)\sin(\theta(t))\cos(\varphi(t))
\end{bmatrix}
\label{eq:af_coordinate_system}
\end{equation}
The time derivatives of the coordinates, expressed in the sensor reference frame, of a stationary reflector
are given by
\begin{equation}
\frac{d\vec{x}}{dt} = -\vec{v}_{s} - \vec{\omega}_{s}\times\vec{x}
\label{eq:af_reflector_time_derivative}
\end{equation}
with $\vec{v}_{s}$ and $\vec{\omega}_{s}$ the linear and rotational velocity vectors respectively of the sensor, expressed in the sensor body reference frame, and $\times$ denoting the cross product. \\
If we choose the x-axis of the mobile platform body reference frame to coincide with the direction of the linear velocity vector and limit the mobile platform to planar movement, i.e. rotations about its z-axis only, the velocity vectors of the sensor are given by
\begin{equation}
\vec{v_{s}} =
\begin{bmatrix}
    V\cos(\alpha + \beta) + \sin(\beta)\cdot l  \omega \\
    V\sin(\alpha + \beta) + \cos(\beta)\cdot l \omega\\
    0
\end{bmatrix},
\vec{\omega_{s}} =
\begin{bmatrix}
    0  \\
    0  \\
    \omega
\end{bmatrix} 
\label{eq:af_velocities}
\end{equation}
\begin{equation}
\delta = \alpha + \beta
\label{eq:delta_ef}
\end{equation}
with $V$ and $\omega$ the magnitudes of the linear and angular velocities of the mobile platform respectively. We will assume these to be constant for the duration of a single sonar measurement. $\alpha$ is the rotation of the sonar sensor around the z-axis of the mobile-platform relative to its linear velocity vector (x-axis), $\beta$ defining the local rotation on the sensor around its local pivot and $l$ the distance between the sensor and the centre of the mobile platform. A visual explanation of these parameters is shown in Fig. \ref{fig:robot_coordinates}. Next we take the derivative of Eq. \ref{eq:af_coordinate_system} 
\begin{equation}
\frac{d\vec{x}}{dt} = 
\begin{bmatrix}
    c\theta & -rs\theta & 0 \\
    -s\theta s\varphi & -rc\theta s\varphi & -rs\theta c\varphi\\
    s\theta c\varphi & rc\theta c\varphi & -rs\theta s\varphi
\end{bmatrix} 
\cdot
\begin{bmatrix}
    dr/dt  \\
    d\theta/dt  \\
    d\varphi/dt
\end{bmatrix} 
\label{eq:af_coordinate_derivative}
\end{equation}
with $r, c\theta, s\theta, c\varphi, s\varphi$ denoting
$r(t), \cos(\theta(t)),\sin(\theta(t)),$ $\cos(\varphi(t)), \sin(\varphi(t))$ respectively. Subsequently, substituting Equations \ref{eq:af_coordinate_system}, \ref{eq:af_velocities} and \ref{eq:af_coordinate_derivative} in Eq. \ref{eq:af_reflector_time_derivative} and restricting the reflectors in the horizontal plane, i.e. $\varphi = \{-\pi/2,\pi/2\}$,  followed by applying trigonometric sum and difference identities, we can express the 2D-velocity field due to mobile platform ego-motion as
\begin{equation}
\begin{bmatrix}
dr/dt\\
d\theta/dt
\end{bmatrix}=\begin{bmatrix}
    -l\omega \sin(\beta - \theta(t)) -V\cos(\theta(t) + \delta)\\
   \cfrac{l\omega \cos(\beta - \theta(t)) + V\sin(\theta(t) + \delta)}{r(t)} + \omega
\end{bmatrix} 
\label{eq:af_refl_ego_motion}
\end{equation}
Do note that the applied restriction to $\varphi$ isn't necessary. Because the autonomous navigation controller used in this paper will only use 2D eRTIS sensor readings, 3D-velocity fields are not required and we can simplify the formulae further by restricting them to the 2D horizontal plane.
\subsection{Linear movement}
In a first case, we restrict the mobile platform to linear movement, i.e. $\omega$ = 0 rad/sec. If we look at the time derivatives of the coordinates $(r(t), \theta(t))$ to construct a 2D acoustic flow model for linear movement, Eq. \ref{eq:af_refl_ego_motion} now reduces to
\begin{equation}
\begin{gathered}
dr/dt = -V\cdot\cos(\theta(t) + \delta) \\
d\theta/dt = \cfrac{V}{r(t)}\cdot\sin(\theta(t) + \delta)
\end{gathered}
\label{eq:af_diff_refl_linear}
\end{equation}
The solution to this set of differential equations has to comply with 
\begin{equation}
\cfrac{\dot{r}(t)}{r(t)} = - \dot{\theta}(t)\cdot\cfrac{\cos(\theta(t) + \delta)}{\sin(\theta(t) + \delta)}
\label{eq:af_diff_refl_linear_comply}
\end{equation}
for $\theta_0 \neq 0$ with $\theta_0$ defined as the angle between the sensor and the line of sight to a particular reflector at its first
sighting. Equivalently, when taking the integral of both sides of Eq. \ref{eq:af_diff_refl_linear_comply} we can define
\begin{equation}
C = |r(t)| \cdot \sin(\theta(t) +\delta)
\label{eq:af_diff_refl_linear_solution}
\end{equation}
This expression describes the invariant for linear mobile platform motions for Multi-Sonar defining the different flow-lines, i.e. the solutions of the differential Eq. \ref{eq:af_diff_refl_linear} for different initial conditions $(r_0, \theta_0)$. Fig. \ref{fig:af_flow_lines_linear} shows example flow-lines for linear movement.
\subsection{Rotation movement}
In the case of a pure rotation of the mobile platform, i.e. $V$ = 0 m/sec, Eq. \ref{eq:af_refl_ego_motion} now reduces the time derivatives of the coordinates $(r(t), \theta(t))$ for the 2D acoustic flow model for rotation movement to 
\begin{equation}
\begin{gathered}
dr/dt = -l\omega \cdot \sin(\beta - \theta(t))\\
d\theta/dt = \cfrac{l\omega}{r(t)} \cdot \cos(\beta - \theta(t)) + \omega
\end{gathered}
\label{eq:af_diff_refl_rotation}
\end{equation}
Fig. \ref{fig:af_flow_lines_rotation} shows example flow-lines for rotation movement. For rotation movement we did not find a closed form expression as we did for linear movement.
\subsection{Other movement}
In the case of the mobile platform following a curved path both a rotation and a translation of the mobile platform will occur. In that case the final flow field will be a combination of the two flow fields described above and can be deduced from Eq. \ref{eq:af_refl_ego_motion}.
\begin{figure}[!ht]
\centering
\includegraphics[width=1\linewidth]{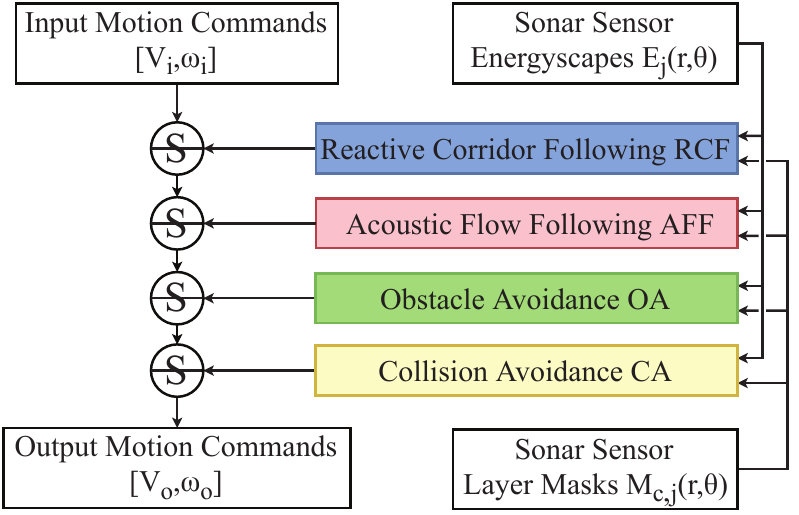}
\caption{Overview of the subsumption architecture of the layered control system used for autonomous navigation. $j$ defines the index of the sensor while $c$ stands for the layer which can be $AFF, RCF, OA$ or $CA$ respectively depending which is used.}
\label{fig:control_layers}
\end{figure}
\section{Layered Control System}\label{sec:control}
The time derivatives for the coordinates of any reflector seen in Section \ref{sec:acoustic_flow_model} can be used to define control laws for autonomous navigation. We can mark out regions around the mobile platform so when a reflection is observed in these regions, certain navigation behaviours are actuated. For defining and activating the various behaviours we use a layered control system inspired by the subsumption architecture proposed by R. A. Brooks \cite{Brooks1986} and for which a schematic is shown in Fig. \ref{fig:control_layers}. The system requires input velocities (linear and rotational $[V_i,\omega_i]$) for the mobile platform which can be created by a manual operator or defined by a global navigation and path finding system. These input motion commands are subsequently modulated by the control behaviour laws. The previous control system is described in detail in \cite{Steckel2017} and has been altered in this paper to be adaptive and stable for Multi-Sonar configurations.
 \begin{figure*}
 \centering
  \begin{tabular}[b]{cc}
    \begin{subfigure}[b]{0.3\linewidth}
      \includegraphics[width=\textwidth]{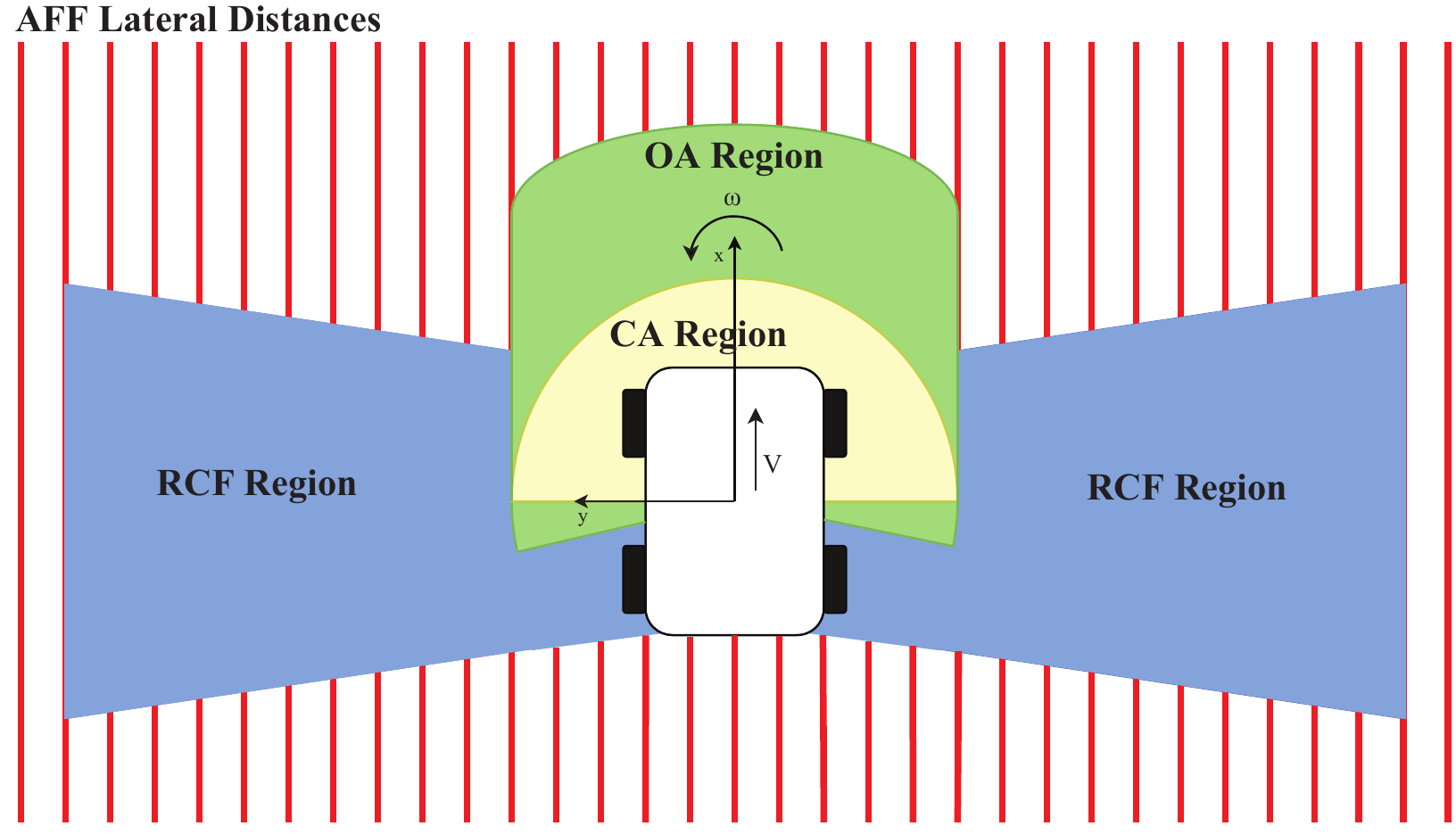}
      \caption{}
      \label{fig:control_robot_regions}
    \end{subfigure}
    &
    \begin{subfigure}[b]{0.65\linewidth}
      \includegraphics[width=\textwidth]{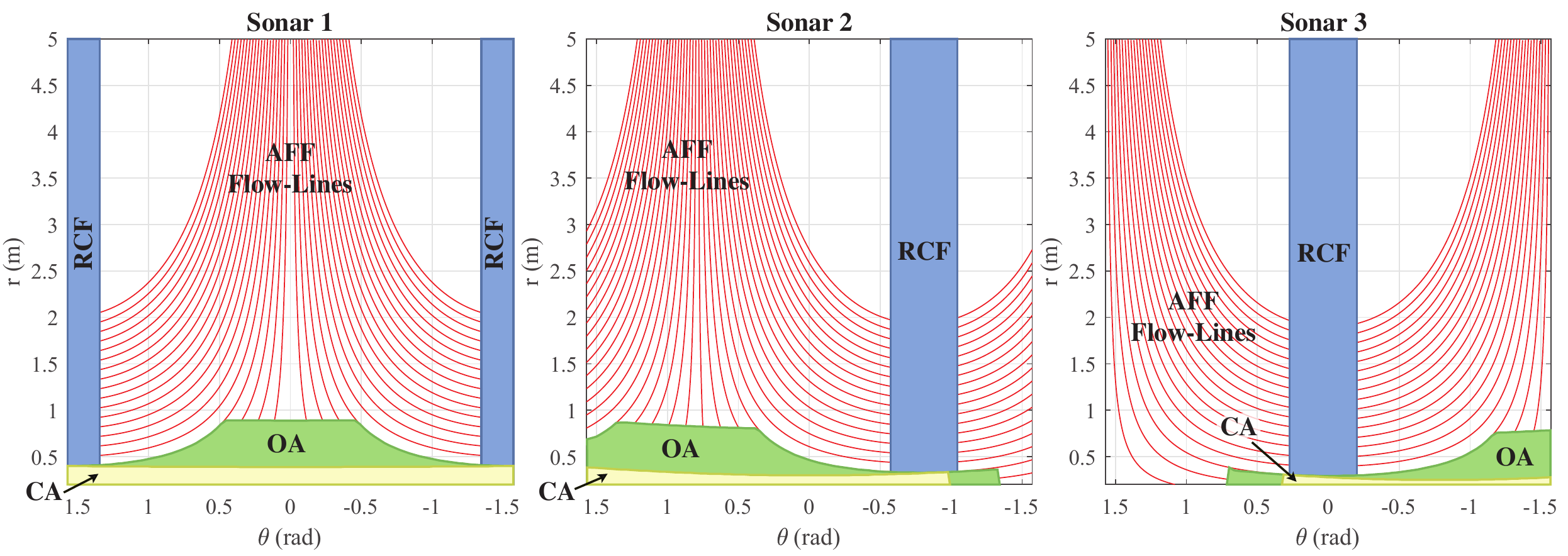}
      \caption{}
      \label{fig:control_masks}
    \end{subfigure}
  \end{tabular}\\
\caption{(a) Each control layer is defined as its spatial region around the mobile platform. These regions can be described by either vertical lines parallel to the mobile platform x-axis, a radius from the mobile platform centre or bounded by specific angles $\theta$ from the centre of the mobile platform. Using these primitive shapes enables the usage of the acoustic flow models to mask these regions within the sensing image for each sensor. The lateral distances used for the Acoustic Flow Following behaviour are also shown; (b) For each sensor as configured and shown in Fig. \ref{fig:af_flow_lines_robot_config} the generated mask is shown for the control regions as defined in Fig. \ref{fig:control_robot_regions}. The flow-lines used for the Acoustic Flow Following behaviour are also shown.}
\label{fig:control}
\end{figure*}   
\subsection{Acoustic control regions}
Each control layer has its spatial representation defined by a region around the mobile platform as can be seen in Fig. \ref{fig:control_robot_regions}. The regions are defined by primitive shapes such as circles, rectangles or trapezoids for the sake of allowing the time differential equations of Section \ref{sec:acoustic_flow_model} to be used to mask these regions within the sensing FOV of each sensor using flow-lines. This way we can create a mask $M_{c,j}$ with $c$ defining the control layer and $j$ the index of the sonar sensor. \\
If we apply these masks to an Energyscape $E_j$ and a reflection is subsequently observed within the masked area, a specific layer can be activated. To allow for the layers to know if a reflection is on the left or right side of the mobile platform (so the layers can decide the correct steering direction), we propose to use ternary masks that are 0 when the voxel is not within the region, and 1 or -1 if the reflection is within the region and respectively on the left or right side. Examples of these masks are shown in Fig. \ref{fig:control_masks}.  
\subsection{Collision avoidance}
The lowest layer with the smallest complexity is Collision Avoidance (CA). This region is activated when the CA-region contains a reflection with a greater intensity than the threshold $T_{CA}$ of the layer. This region as seen on Fig. \ref{fig:control_robot_regions} is masked for each sensor by its generated $M_{CA,j}$ as illustrated on Fig. \ref{fig:control_masks}.\\ The activation control law $C_{CA}$ is defined as 
\begin{equation}
C_{CA} : \exists r,\theta: \sum_{r,\theta,j} E_j(r,\theta)\cdot M_{CA,j}(r,\theta) > T_{CA}
\label{eq:ca_law}
\end{equation}
The robot will rotate in the opposite direction of the closest reflector at a rotational output velocity $\omega_o=$ set at a fixed rate. In our simulations we used \SI{0.5}{\radian\per\second}. The direction of the closest reflection is known by using the ternary masking. Subsequently, calculating the total sum of the multiplication of the mask and Energyscapes has a positive or negative value causing steering left or right respectively.
\newpage 
\noindent The linear output velocity $V_o$ is \SI{0}{\meter\per\second} until CA is activated consecutively for four times or more. This value eliminates most false-positives and worked well in our simulation. In that case the linear output velocity is set to \SI{-0.1}{\meter\per\second} so it reverses slightly. These velocities will be maintained until the CA-region is cleared of any reflection greater than $T_{CA}$. 
\subsection{Obstacle avoidance}
The second layer is Obstacle Avoidance (OA). This region is activated when a reflection in the masked Energyscape is larger than the threshold value $T_{OA}$. The region, which can be seen in Fig. \ref{fig:control_robot_regions} is created such that when a linear velocity $V$ is maintained forward (with $\omega=0$), a collision will happen. As such this layer will avoid this collision with the control law $C_{OA}$
\begin{equation}
\begin{split}
& C_{OA} : \exists r,\theta: E_j(r,\theta)\cdot M_{OA,j}(r,\theta) > T_{OA}\\
& \omega_{OA} = \omega_i + \lambda_{OA}\cdot \cfrac{\sum_{r,\theta,j} \frac{1}{r^2} \cdot E_j(r,\theta)\cdot M_{OA,j}(r,\theta)}{\sum_{r,\theta,j}E_j(r,\theta)\cdot M_{OA,j}(r,\theta)}\\
& V_{OA} = V_i\cdot\left[1-\mu_{OA}\cdot \sum_{r,\theta,j}E_j(r,\theta)\cdot M_{OA,j}(r,\theta)\right]
\end{split}
\label{eq:oa_law}
\end{equation}
This layer will calculate the output velocities $V_o=V_{OA}$ and $\omega_o=\omega_{OA}$ and steer away from the side of the mobile platform with the greatest total reflection within the masked region. By using the ternary masks the sign of the rotation (left or right) will be set correctly. By dividing the sum of the masks and the Energyscapes with the square of the range of each reflection, further away reflections will cause less avoidance motion. Furthermore, this layer will lower the linear output velocity depending on the total reflection. Gain factors $\lambda_{OA}$ and $\mu_{OA}$ are always set to 1 in our results but can be altered to fine-tune the behaviour.
\subsection{Reactive corridor following}
Reactive Corridor Following (RCF) will create initial alignment with a corridor quickly. This is done intentionally so that Acoustic Flow Following (AFF) can take over. \\It does so by equalising the total reflection in the Energyscape in the left and right peripheral regions, i.e. by steering away from the side with the most energy in the fused masked Energyscapes of the sonars. This region can be observed in Fig. \ref{fig:control_robot_regions} while the corresponding masks can be observed in Fig. \ref{fig:control_masks}. It will be activated once a reflection in these zones is greater than the threshold $T_{RCF}$.\\
The law $C_{RCF}$ is expressed as
\begin{equation}
\begin{split}
& C_{RCF} : \exists r,\theta: E_j(r,\theta)\cdot M_{RCF,j}(r,\theta) > T_{RCF}\\
& \omega_{RCF} = \omega_i + \lambda_{RCF}\cdot \cfrac{\sum_{r,\theta,j} \frac{1}{r^2} \cdot E_j(r,\theta)\cdot M_{RCF,j}(r,\theta)}{\sum_{r,\theta,j}E_j(r,\theta)\cdot M_{RCF,j}(r,\theta)}\\
& V_{RCF} = V_i
\end{split}
\label{eq:rcf_law}
\end{equation}
This layer will calculate the output velocities $V_o=V_{RCF}=V_i$ and $\omega_o=\omega_{RCF}$ with gain factor $\lambda_{RCF}$ set to 1 in our results.
\subsection{Acoustic flow following}
When alignment with a single wall or a corridor has been achieved by the RCF-layer or otherwise, the AFF layer is activated. If a mobile platform is aligned with one or multiple parallel walls all reflections on the wall(s) will exist on the same flow-line defined by the acoustic flow model for linear movement as seen in Section \ref{sec:acoustic_flow_model} and on Fig. \ref{fig:af_flow_lines_linear}. We can measure that alignment for a sensor $j$ with 
\begin{equation}
\Gamma_j(d) = \cfrac{\sum_{r,\theta}E_j(r,\theta)\cdot M_{AFF,F_{d,j}}(r,\theta) \cdot  \sqrt[\leftroot{0}\uproot{3}]{r}}{|F_{d,j}|}
\label{eq:AF_integral}
\end{equation}
with $F_{d,j}$ representing the flow-line associated with the reflections of the features on those walls at a lateral distance $d$ from the mobile platform path and $|F_{d,j}|$ denoting its length, i.e. the number of voxels lying on that flow-line. This formula can be seen as calculating the integral of that flow-line. \\
We can calculate $F_{d,j}$ for each sensor and each lateral distance $d$ defined by the control system. Examples of these distances that can be chosen can be seen in Fig. \ref{fig:control_robot_regions} with corresponding flow-lines shown in Fig. \ref{fig:control_masks}. Afterwards, fusing all alignment values from each sensor gives the total $\Gamma(d)$. 
\begin{table*}
\centering
\caption{Simulation Configurations}
\label{table:results_configurations}
\resizebox{1.42\columnwidth}{!}{%
\begin{tabular}{@{}c|ccc|ccc|ccc@{}}
\multirow{2}{*}{Setup} & \multicolumn{3}{c|}{eRTIS 1} & \multicolumn{3}{c|}{eRTIS 2} & \multicolumn{3}{c}{eRTIS 3} \\ \cline{2-10}
\addlinespace[0.1em]
& $\alpha$ & $\beta$ & $l$ & $\alpha$ & $\beta$ & $l$ & $\alpha$ & $\beta$ & $l$ \\ \hline
\addlinespace[0.1em]
1  & \ang{0} & \ang{0} & \SI{18}{\cm} &  & unused &   &   & unused &   \\ 
2  & \ang{0} & \ang{-20} & \SI{14}{\cm}  &\ang{90} & \ang{-10} & \SI{10}{\cm}  & \ang{-90} & \ang{-5} & \SI{8}{\cm}  \\ 
3  & \ang{90} & \ang{-20} & \SI{10}{\cm}  &\ang{-90} & \ang{20} & \SI{10}{\cm}  & & unused &\\ 
4  & \ang{0} & \ang{0} & \SI{12}{\cm} &\ang{90} & \ang{0} & \SI{12}{\cm}  & \ang{-90} & \ang{0} & \SI{12}{\cm}  \\ 
5  & \ang{45} & \ang{0} & \SI{4}{\cm}  &\ang{-135} & \ang{0} & \SI{4}{\cm}  &  & unused &  \\ 
6  & \ang{0} & \ang{0} & \SI{10}{\cm}  &\ang{-180} & \ang{0} & \SI{0}{\cm}  &   & unused &  \\ 
7  & \ang{0} & \ang{20} &\SI{ 6}{\cm}  &\ang{90} & \ang{10} & \SI{0}{\cm}  & \ang{-90} & \ang{20} & \SI{14}{\cm}  \\ 
8  & \ang{0} & \ang{0} & \SI{0}{\cm}  &\ang{120} & \ang{-120} & \SI{14}{\cm}  & \ang{-120} & \ang{120} & \SI{14}{\cm}  \\ 
9  & \ang{180} & \ang{-180} &\SI{6}{\cm}  &  & unused &   &   & unused & \\ 
10 & \ang{45} & \ang{-10} & \SI{6}{\cm}  &\ang{-45} & \ang{10} & \SI{6}{\cm}  & \ang{-180} & \ang{0} & \SI{0}{\cm}  \\ 
\end{tabular}%
}
\end{table*}
\newpage
\noindent The AFF-layer is activated when at least one $\Gamma(d)$ value peaks and is greater than the threshold $T_{AF,single}$. In that case the layer will cause single wall following behaviour for that peak at lateral distance $d_{s}$ defined as 
\begin{equation}
\begin{split}
& \omega_{AFF} = 
\begin{cases}
      \omega_i + \lambda_{AFF} \cdot( d_s - d_p) & \text{if } d_p \text{ is defined}\\
      \omega_i & \text{otherwise}
\end{cases} \\
& V_{AFF} = V_i
\end{split}
\label{eq:AF_single_law}
\end{equation}
with $d_p$ being the previous lateral distance at which the wall was detected. Therefore, the single wall AFF behaviour only results in an added rotation if the layer is activated consecutively twice or more.\\
Corridor following behaviour is activated if two peaks at lateral distances $d_l$ and $d_r$ are greater than the threshold $T_{AFF,corr}$ and are on opposite sides of the mobile platform, i.e. on opposite sides of $d=0$. The resulting behaviour is defined as
\begin{equation}
\begin{split}
& \omega_{AFF} = \omega_i + \lambda_{AFF}\cdot (d_l - d_r)\\
& V_{AFF} = V_i
\end{split}
\label{eq:AF_corridor_law}
\end{equation}
with gain factor $\lambda_{AFF}$ for both single wall and corridor following behaviour being set to 1 in our simulations.
\section{Experimental Results}\label{sec:results}
\begin{figure*}[!ht]
\centering
\includegraphics[width=0.8\linewidth]{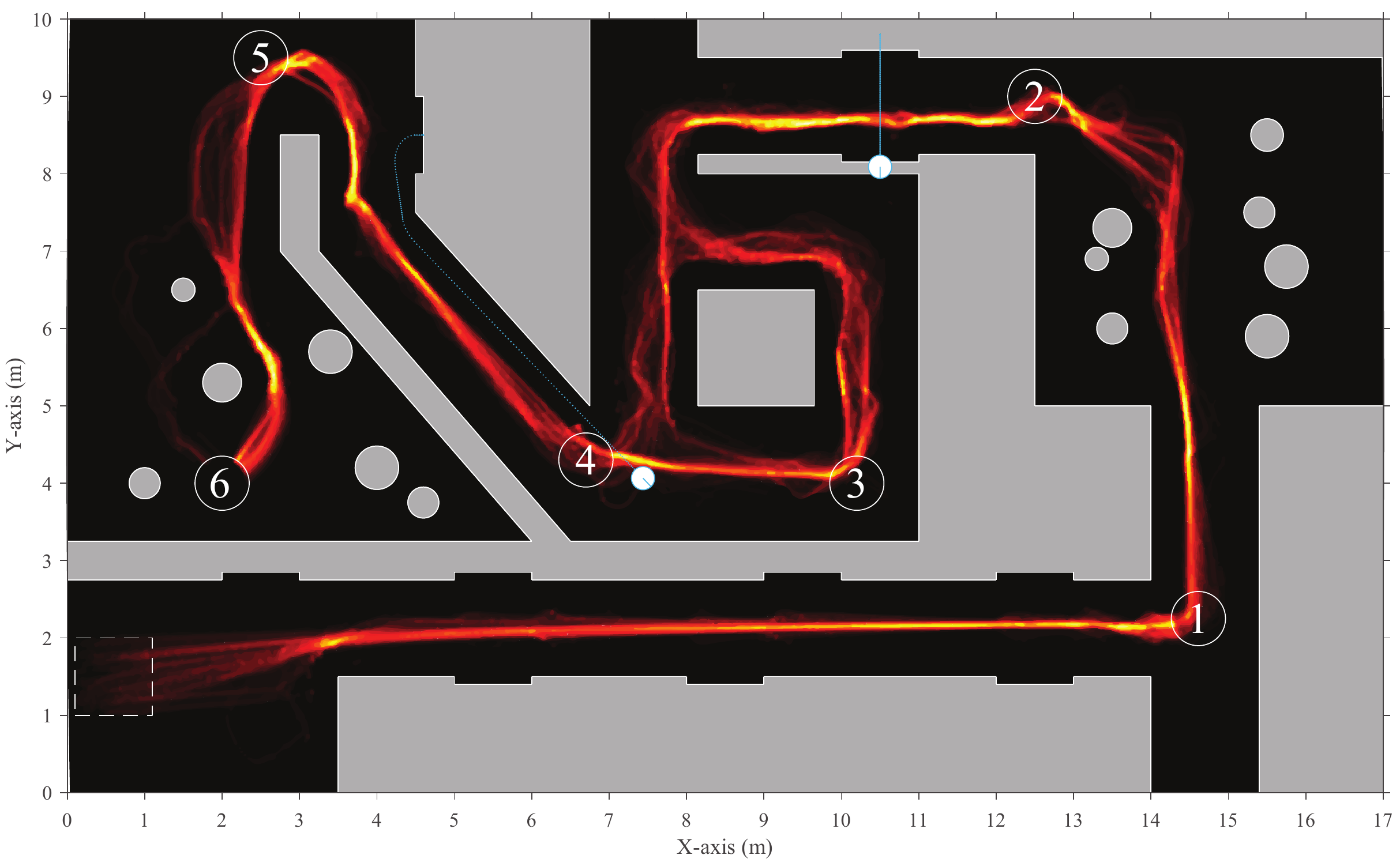}
\caption{A heat map of the trajectory distribution of a small robot autonomously navigating through a environment in simulation. The distribution was taken over 10 different Multi-Sonar configurations with each being run 15 times. The start zone (rectangle with dashed edges) is where within a random start position is chosen for each run. After that it will move between the numbered waypoints.}
\label{fig:results_heat_map}
\end{figure*}

For validating that the proposed acoustic flow-based layered control system works, we developed a simulation of the world, mobile platform and eRTIS sensors. This allows us to test this system and perform a numerical evaluation safely before we can make a complete analysis with real world validation. In \cite{Steckel2017} a previous, single sensor control system was created in the same simulator, and deployed this developed control law on a real-world robot, indicating the validity of the used simulation paradigm used here. For the results of this paper we tried to see if the mobile platform could navigate several environments without colliding with other objects and show stable behaviour independent of the configurations of the Multi-Sonar sensors. These environments were comprised of corridors with doors, single walls, circular objects and a set of moving objects representing potential humans or other mobile platforms. This last category resembles for example other AGVs or humans traversing the environment.
In the simulation we used a small differential steering robot with a \SI{20}{\cm} radius that travels up to \SI{0.3}{\meter\per\second}. \\ 
The control system always requires input velocities as seen in Section \ref{sec:control}. For these simulation experiments, we used a waypoint guidance navigation system based on the work proposed by P. Boucher \cite{Boucher2016} which is designed specifically for differential steering vehicles. These waypoints are placed sparsely to force the control system to figure out its own local path between each waypoint. \\
For simulating the sonar sensor we create impulse responses for the most relevant acoustic reflection types such as corners, edges and planar surfaces \cite{Kuc1987}, with added noise and includes proper sound occlusion of objects. Upon that, we then use these responses as input for our time-domain simulation of the eRTIS sensor where we use 32 virtual microphone signals to simulate the sensory recording data. We post-process this data and create 2D Energyscapes of the horizontal plane in front of the sensor (azimuth ranging from \ang{-90} to \ang{90} in steps of \ang{1}) with a range of \SI{5}{\meter}. \\
For validation of the adaptive nature of the new control system for Multi-Sonar we tested 10 different Multi-Sonar configurations, ranging from one to three sonar sensors. These configurations are listed in Table \ref{table:results_configurations}. Each configuration was ran through the same environment 15 times. One simulation environment and its results is shown in Fig. \ref{fig:results_heat_map}. As seen on that figure and from our own analysis of this and other environments we tested, not a single configuration tested caused a collision or extreme behaviour that caused it to go off the intended path. We see the expected behaviour of following corridors or walls, navigating junctions, avoiding collisions with stationary or dynamic objects as well as being drawn to the waypoints by the global navigation system. The main differences that can be observed are in the choice of the safest path based on the sensor readings, which will differ between Multi-Sonar configurations. Finally, we observed that by using a sparse set of waypoints, the intended behaviour of the layers was often slightly disturbed due to being too strongly drawn to the waypoints. However, this is not a fault of the control system but of the generated input velocities. This could be solved by using a different global path strategy or increase the amount of waypoints. We did not experiment using only part of the control system by deactivating certain layers as they have been individually tested and demonstrated in \cite{Steckel2017} extensively. 
\section{Conclusions \& Future Work}\label{sec:conclusion}
The results demonstrate that an autonomous mobile platform can navigate continuously through spatially varied and dynamic environments with one or more 3D-sonar sensors in simulation. Allowing these sensors to be used in optimally fitted configurations for various vehicle types. We could not observe any deadlock behaviour or collisions in our results. However, specific experiments could be performed in the future to completely ruled these out. \\
This system accomplishes spatial navigation without explicit spatial segmentation. Furthermore, by creating a modular and adaptive approach for the creation of both the acoustic flow model and the layered control system for Multi-Sonar we can use this system on any mobile platform, independent of the configuration of the sensors. The current layers haven proven to be sufficient for safe navigation between targets but more layers could be added for more complex behaviour, e.g. when near specific objects such as for example charging stations, doors or elevators. \\
With the numerical evaluation shown in our results and the theoretical foundation completed we can now implement an online real-time system on a real-world mobile platform and perform a full analysis of the navigation system. Furthermore, the task ahead requires further research on using 3D-sonar sensors to assist autonomous agents in completing their tasks. One such task is creating a method that could be used to provide a comprehensive solution for path planning, task scheduling and global navigation. So that a more optimal global path can be created. Thereupon, a system such as presented in this paper could serve for the local navigation. However, to accomplish this task more spatial information of the environment is needed than found in the Energyscapes. This spatial information could be created for example by sonar SLAM algorithms such as \cite{Steckel2013b, Krekovic2016} or using beacons\cite{Simon2020}. However, similarly as with the systems provided in this paper, these solutions need to be made modular and adaptive to any Multi-Sonar configuration.
\bibliographystyle{IEEEtran}
\bibliography{toot}
\end{document}